\documentclass{article}

\usepackage{microtype}
\usepackage{graphicx}
\usepackage{subfigure}
\usepackage{booktabs} 

\usepackage{hyperref}

\usepackage[accepted]{icml2024}

\usepackage{amsmath}
\usepackage{amssymb}
\usepackage{mathtools}
\usepackage{amsthm}

\usepackage[capitalize,noabbrev]{cleveref}

\theoremstyle{plain}

\theoremstyle{definition}

\theoremstyle{remark}

\usepackage[textsize=tiny]{todonotes}

\icmltitlerunning{The Missing Curve Detectors of InceptionV1}

\begin{document}

\twocolumn[
\icmltitle{The Missing Curve Detectors of InceptionV1: Applying Sparse Autoencoders to InceptionV1 Early Vision}



\icmlsetsymbol{equal}{*}

\begin{icmlauthorlist}
\icmlauthor{Liv Gorton}{yyy}
\end{icmlauthorlist}

\icmlaffiliation{yyy}{Cellware Labs, San Francisco, CA, United States of America}

\icmlcorrespondingauthor{Liv Gorton}{liv@livgorton.com}

\icmlkeywords{mechanistic interpretability, InceptionV1, sparse autoencoders, superposition, vision interpretability}

\vskip 0.3in
]



\printAffiliationsAndNotice{} 

\begin{abstract}
Recent work on sparse autoencoders (SAEs) has shown promise in extracting interpretable features from neural networks and addressing challenges with polysemantic neurons caused by superposition. In this paper, we apply SAEs to the early vision layers of InceptionV1, a well-studied convolutional neural network, with a focus on curve detectors. Our results demonstrate that SAEs can uncover new interpretable features not apparent from examining individual neurons, including additional curve detectors that fill in previous gaps. We also find that SAEs can decompose some polysemantic neurons into more monosemantic constituent features. These findings suggest SAEs are a valuable tool for understanding InceptionV1, and convolutional neural networks more generally.

\end{abstract}

\section{Introduction}

The original project of mechanistic interpretability was a thread of work \cite{cammarata2020thread} trying to reverse engineer InceptionV1 \cite{DBLP:journals/corr/SzegedyLJSRAEVR14}. However, this work had a major limitation that was recognised at the time: polysemantic neurons which respond to unrelated stimuli. It was hypothesised that this was due to \emph{superposition}, where combinations of neurons are used to represent features \cite{arora2018linear,elhage2022toy}. This was a significant barrier to the original Circuits project, since uninterpretable polysemantic neurons were a roadblock for neuron-based circuit analysis.

Since then, significant progress has been made on addressing superposition. In particular, it has been found that applying a variant of dictionary learning \cite{olshausen1997sparse,elad2010sparse} called a sparse autoencoder (SAE) can extract interpretable features from language models \cite{cunningham2023sparse,yun2021transformer,bricken2023monosemanticity}.

This creates an opportunity for us to return to the original project of understanding InceptionV1 with new tools. In this paper, we apply sparse autoencoders to InceptionV1 early vision \cite{olah2020an} and especially curve detectors \cite{cammarata2020curve,cammarata2021curve}. We find that at least some previously uninterpretable neurons contribute to representing interpretable features. We also find ``missing features" which are invisible when examining InceptionV1 in terms of neurons but are revealed by features. This includes a number of additional curve detector features which were missing from the well-studied neuron family.


\section{Methods}

\subsection{Sparse Autoencoders}

Following previous work, we decompose any activation vector $x$ into $x \simeq b + \sum_if_i(x)d_i$ where $f_i(x)$ is the feature activation and $d_i$ is the feature direction. Throughout the paper, we'll denote these features as \texttt{<layer>/f/<i>} (and neurons as \texttt{<layer>/n/<i>}).

\begin{table}[t]
\vskip 0.15in
\begin{center}
\begin{small}
\begin{tabular}{lcccr}
\toprule
Layer & $\lambda$ & Expansion Factor \\
\midrule
conv2d0    & $1\text{e}-7$ & $2\times$ & \\
conv2d1 & $5\text{e}-6$& $4\times$ &\\
conv2d2    & $5\text{e}-6$& $4\times$ &\\
mixed3a    & $3.5\text{e}-6$& $8\times$ &\\
mixed3b     & $3.5\text{e}-6$& $8\times$ &\\
\bottomrule
\end{tabular}
\end{small}
\end{center}
\caption{Default sparse autoencoder hyperparameters for each layer.}
\label{table:run-overview}
\vskip -0.1in
\end{table}

\begin{figure*}[hbt!]
\begin{center}
\centerline{\includegraphics[width=\textwidth]{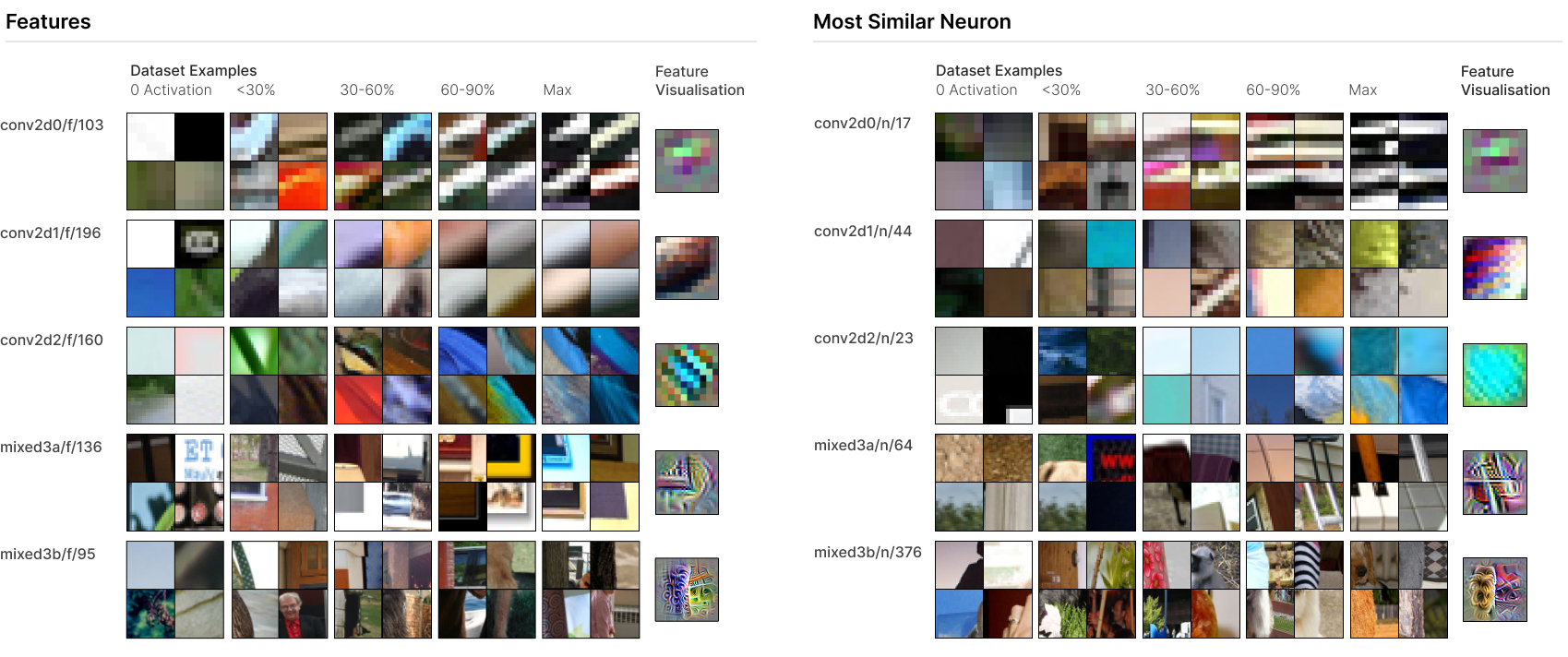}}
\caption{This figure presents examples of one interpretable feature learned by the SAE at each layer. For each feature, we present dataset examples at varying activation levels and a feature visualisation. We also show the neuron most similar to this feature, and corresponding dataset examples and feature visualisations.}
\label{figure:example-features}
\end{center}
\end{figure*}

We achieve this decomposition using SAEs trained on image activations sampled from InceptionV1 over its training set, ImageNet Large Scale Visual Recognition Challenge (ILSVRC) \cite{imagenet-object-localization-challenge}. Recall that, as a conv net, InceptionV1 has a grid of activation vectors corresponding to image positions. For each of the images in ILSVRC, we sample activations from 10 such positions. We trained for approximately 500 epochs on a dataset of shuffled activations (until a total of 5 billion activations were trained on).

Unless otherwise stated, the hyperparameters used for each SAE can be found in \cref{table:run-overview}. Our specific SAE setup followed some modifications recommended by \citet{conerly2024update}. In particular, we didn't constrain the decoder norm and instead scaled feature activations by it.\footnote{This seemed to help avoid dead neurons.} We also made two additional modifications to work around compute constraints, described in the following subsections.

\subsubsection{Over Sampling Large Activations}

In InceptionV1, the majority of image positions produce small activations (for example, on backgrounds). To avoid spending lots of compute modelling these small activations, we oversampled large activations by sampling activations from positions in the image proportional to their activation magnitude.

\subsubsection{Branch Specific SAEs}

InceptionV1 divides these layers into branches with different convolution sizes \cite{DBLP:journals/corr/SzegedyLJSRAEVR14}. For layers \texttt{mixed3a} and \texttt{mixed3b}, we perform dictionary learning only on the \texttt{3x3} and \texttt{5x5} convolutional branches respectively. This seems relatively principled as with such limited space, features are likely isolated to a branch that has the most advantageous filter size. (Since the original publication of this work, we have performed experiments that revealed cross-branch superposition is much more significant than our intuition suggested. Further information can be found in \ref{branch-specialisation-appendix} and this doesn't undermine the correctness of the results presented in this paper.)

\subsection{Analysis Methods}

In analysing features, we primarily rely on \emph{dataset examples} and \emph{feature visualisation} \cite{erhan2009visualizing,olah2017feature}. Dataset examples show how a neuron or feature behaves on distribution, while feature visualisation helps isolate what causes a feature to activate. This is supplemented by more specific methods like synthetic dataset examples from \citet{cammarata2020curve} for curve detectors.

\subsubsection{Dataset Examples}

We collect dataset examples where the feature activates over the ILSVRC dataset \cite{imagenet-object-localization-challenge}, since that is what InceptionV1 was trained on. In using dataset examples, it's important to not just look at top examples to avoid interpretability illusions \cite{bolukbasi2021interpretability}. Instead, we collect dataset examples randomly sampled to be within ten activation intervals, evenly spaced between 0 and the feature's maximum observed activation.

\begin{figure*}[hbt!]
\begin{center}
\centerline{\includegraphics[width=\textwidth]{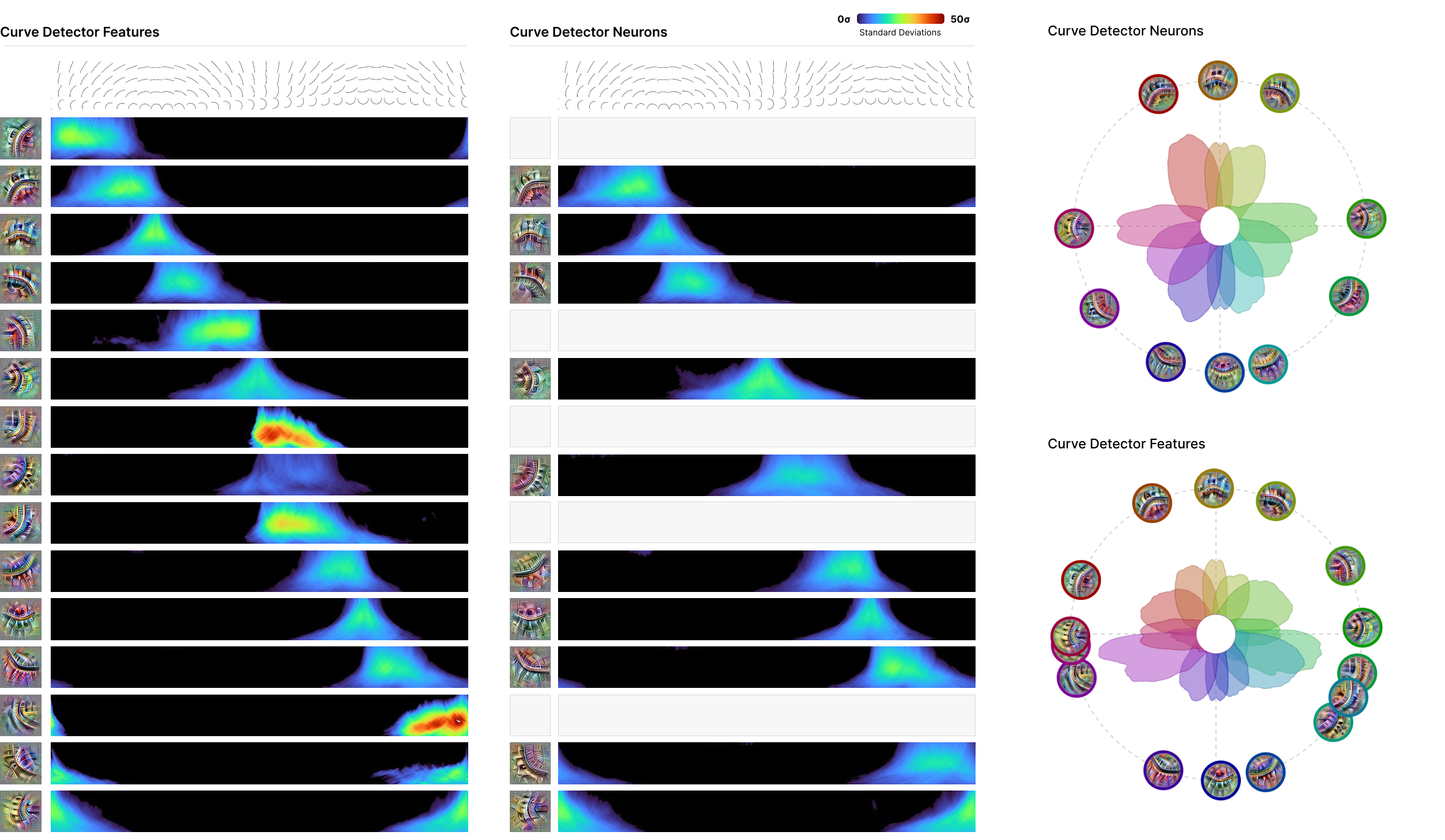}}
\caption{\textbf{Left \& Middle:} Synthetic data plots, showing how curve detector features (left) and neurons (middle) respond to synthetic curve stimuli, as in \citet{cammarata2020curve}. Activations are denominated in standard deviations. A subset of the new curve detectors are shown, with gaps between the neurons representing a previously missing curve detector. \textbf{Right:} The same data is shown on radial tuning curve plots, again following \citet{cammarata2020curve}. The curve radius at a given orientation denotes activation, again measured in standard deviations of activation.}
\label{curve-detector-responses}
\end{center}
\end{figure*}

\subsubsection{Feature Visualisation}

Feature visualisation is performed following \citet{olah2017feature}. We use the Lucent library \cite{torch-lucent}, with some small modifications to their port of InceptionV1 to ensure compatibility with the original model.\footnote{It was important to us to ensure our model was identical to the one studied by the original Circuits thread. In cross-validating the torch version with the original TensorFlow one, we found that some small differences were introduced to the local response normalisation layer when Lucent ported the model to PyTorch. These significantly modify model behaviour. A PR for these fixes can be found at: \href{https://github.com/greentfrapp/lucent/pull/51}{https://github.com/greentfrapp/lucent/pull/51}}

\section{Results}

This paper is a very early report on applying SAEs to InceptionV1. We present a variety of \emph{examples} of SAEs producing more interpretable features in various ways. Our sense is that these results are representative of a more general trend, but we don't aim to defend this in a systematic way at this stage.

Concretely, we claim the following:
\vspace{-4mm}
\begin{enumerate}
  \setlength{\itemsep}{3pt}
  \setlength{\parskip}{0pt}
   \item There exist relatively interpretable SAE features which were not visible in terms of neurons.
   \item There are new curve detector features which fill missing ``gaps" among curve detectors found by \citet{cammarata2020curve}.
   \item Some polysemantic neurons can be seen to decompose into more monosemantic features.
\end{enumerate}

Each of the following subsections will support one of these claims.

\subsection{SAEs Learn New, Interpretable Features}

\cref{figure:example-features} presents five SAE features, one from each ``early vision" layer we applied SAEs to. (These are the same five layers studied by \citet{olah2020an}.) In each case, the feature is quite interpretable across the activation spectrum (see dataset examples in \cref{figure:example-features}). The neurons most involved in representing the feature are, to varying extents, vaguely related or polysemantic neurons, which are difficult to interpret.

For example, \texttt{conv2d1/f/196} appears to be a relatively monosemantic brightness gradient detector. However, the most involved neuron \texttt{conv2d1/n/44} was unable to be categorised by \citet{olah2020an} and appears quite polysemantic. (In fact, our SAE decomposes \texttt{conv2d1/n/44} into a variety of features responding to brightness gradients of various orientations, colour contrast, and complex Gabor filters.) On the other hand, there are features which have more monosemantic corresponding neurons, but are still more precise and monosemantic. For example, \texttt{mixed3b/f/95} is a relatively generic boundary detector receiving its largest weight from \texttt{mixed3b/n/376}, which responds to boundaries with a preference for fur on one side. The SAE appears to decompose \texttt{mixed3b/n/376} into a generic boundary detector and a fur-specific boundary detector in the same orientation.

\subsection{SAEs Discover Additional Curve Detectors}

Curve detectors are likely the best characterised neurons in InceptionV1, studied in detail by \citet{cammarata2020curve,cammarata2021curve}. They were also found to be concentrated within the \texttt{5x5} branch of \texttt{mixed3b} by \citet{voss2021branch}. This existing literature makes them a natural target for understanding the behaviour of SAEs.

In \cref{curve-detector-responses}, we use the ``synthetic data" approach of \citet{cammarata2020curve} to systematically show how our curve detector features respond to curve stimuli of a range of orientations and radii. We find curve detector features which closely match the curve detector neurons known to exist, and others which fill in ``gaps" where there wasn't a curve detector for a particular orientation. Following existing work, we report the stimuli activations in terms of standard deviations to make it clear how unusual the kind of intense reaction they produce is. To see the way the new curve detector features ``fill in" gaps, we also create ``radial tuning curve" plots from \citet{cammarata2020curve}, seen in \cref{curve-detector-responses}. The features fill in several gaps where there is no neuron responding to curves in a given orientation.

\subsection{SAEs Split Polysemantic Neurons}

We've already mentioned several examples of polysemantic neurons being split into more monosemantic features. In this section, we'll further support this pattern with a particularly striking example. \citet{olah2020an} reported the existence of ``double curve" detector neurons, which respond to curves in two very different orientations. It's very natural to suspect these are an example of superposition.

\begin{figure}[ht]
\vskip 0.2in
\begin{center}
\centerline{\includegraphics[width=\columnwidth]{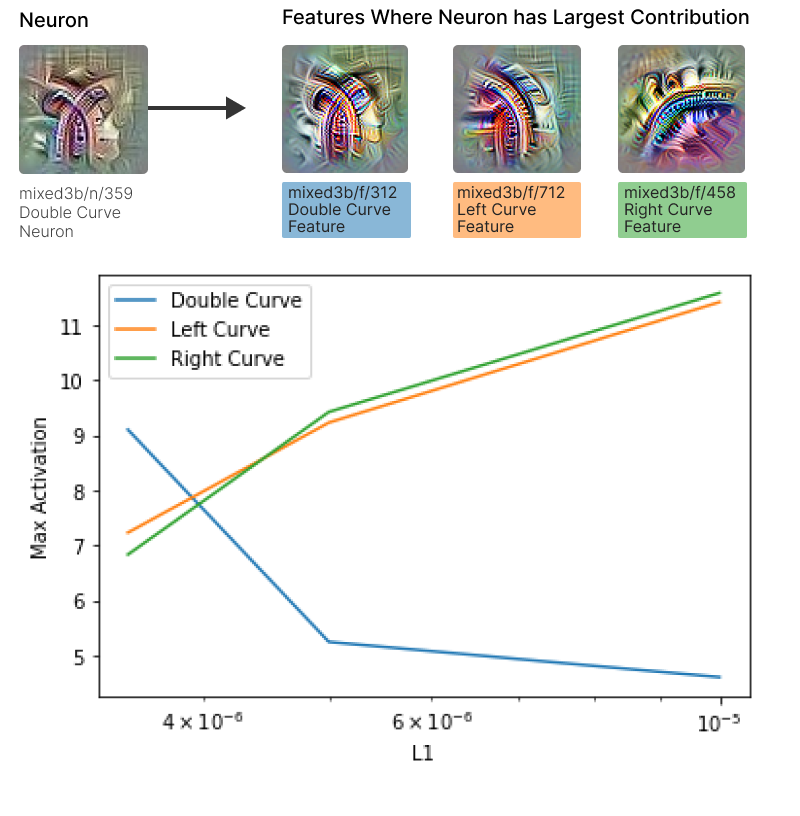}}
\caption{\textbf{Top:} The three SAE features most strongly weighted to \texttt{mixed3b/n/359}, previously identified by \citet{olah2020an} as a double curve detector that was likely polysemantic. \textbf{Bottom:} Max activations of analogous features across SAEs with different L1 coefficients. As L1 increases, the double curve feature becomes smaller, while the left and right curves correspondingly grow.}
\label{double-curve-detector}
\end{center}
\vskip -0.2in
\end{figure}

As shown in \cref{double-curve-detector}, the double curve detector \texttt{mixed3b/n/359} primarily decomposes into three features: two monosemantic curve detectors and a double curve detector. We suspected the additional double curve detector might be a failure of the SAE. To study this, we trained two additional SAEs with higher L1 coefficients. As L1 the coefficient increases, the maximum activation of the analogous double curve features falls, while the left and right curve features maximum activations correspondingly increases (see bottom of \cref{double-curve-detector}).


\section{Conclusion}

Recent progress on SAEs appears to open up exciting new mechanistic interpretability directions in the context of language models, such as circuit analysis based on features \citep{marks2024sparse}. Our results suggest that this promise isn't limited to language models. At the very least, SAEs seem promising for understanding InceptionV1.



\section*{Acknowledgements}

I would like to thank my co-founder, Shae McLaughlin, for her ongoing support. I am also grateful to Chris Olah for insightful discussions and feedback that helped shape this work.

\bibliography{bibliography}

\begin{thebibliography}{20}
\providecommand{\natexlab}[1]{#1}
\providecommand{\url}[1]{\texttt{#1}}
\expandafter\ifx\csname urlstyle\endcsname\relax
  \providecommand{\doi}[1]{doi: #1}\else
  \providecommand{\doi}{doi: \begingroup \urlstyle{rm}\Url}\fi

\bibitem[Addison~Howard(2018)]{imagenet-object-localization-challenge}
Addison~Howard, Eunbyung~Park, W.~K.
\newblock Imagenet object localization challenge, 2018.
\newblock URL \url{https://kaggle.com/competitions/imagenet-object-localization-challenge}.

\bibitem[Arora et~al.(2018)Arora, Li, Liang, Ma, and Risteski]{arora2018linear}
Arora, S., Li, Y., Liang, Y., Ma, T., and Risteski, A.
\newblock Linear algebraic structure of word senses, with applications to polysemy.
\newblock \emph{Transactions of the Association for Computational Linguistics}, 6:\penalty0 483--495, 2018.

\bibitem[Bolukbasi et~al.(2021)Bolukbasi, Pearce, Yuan, Coenen, Reif, Vi{\'e}gas, and Wattenberg]{bolukbasi2021interpretability}
Bolukbasi, T., Pearce, A., Yuan, A., Coenen, A., Reif, E., Vi{\'e}gas, F., and Wattenberg, M.
\newblock An interpretability illusion for bert.
\newblock \emph{arXiv preprint arXiv:2104.07143}, 2021.

\bibitem[Bricken et~al.(2023)Bricken, Templeton, Batson, Chen, Jermyn, Conerly, Turner, Anil, Denison, Askell, Lasenby, Wu, Kravec, Schiefer, Maxwell, Joseph, Hatfield-Dodds, Tamkin, Nguyen, McLean, Burke, Hume, Carter, Henighan, and Olah]{bricken2023monosemanticity}
Bricken, T., Templeton, A., Batson, J., Chen, B., Jermyn, A., Conerly, T., Turner, N., Anil, C., Denison, C., Askell, A., Lasenby, R., Wu, Y., Kravec, S., Schiefer, N., Maxwell, T., Joseph, N., Hatfield-Dodds, Z., Tamkin, A., Nguyen, K., McLean, B., Burke, J.~E., Hume, T., Carter, S., Henighan, T., and Olah, C.
\newblock Towards monosemanticity: Decomposing language models with dictionary learning.
\newblock \emph{Transformer Circuits Thread}, 2023.
\newblock https://transformer-circuits.pub/2023/monosemantic-features/index.html.

\bibitem[Cammarata et~al.(2020{\natexlab{a}})Cammarata, Carter, Goh, Olah, Petrov, Schubert, Voss, Egan, and Lim]{cammarata2020thread}
Cammarata, N., Carter, S., Goh, G., Olah, C., Petrov, M., Schubert, L., Voss, C., Egan, B., and Lim, S.~K.
\newblock Thread: Circuits.
\newblock \emph{Distill}, 2020{\natexlab{a}}.
\newblock \doi{10.23915/distill.00024}.
\newblock https://distill.pub/2020/circuits.

\bibitem[Cammarata et~al.(2020{\natexlab{b}})Cammarata, Goh, Carter, Schubert, Petrov, and Olah]{cammarata2020curve}
Cammarata, N., Goh, G., Carter, S., Schubert, L., Petrov, M., and Olah, C.
\newblock Curve detectors.
\newblock \emph{Distill}, 2020{\natexlab{b}}.
\newblock \doi{10.23915/distill.00024.003}.
\newblock https://distill.pub/2020/circuits/curve-detectors.

\bibitem[Cammarata et~al.(2021)Cammarata, Goh, Carter, Voss, Schubert, and Olah]{cammarata2021curve}
Cammarata, N., Goh, G., Carter, S., Voss, C., Schubert, L., and Olah, C.
\newblock Curve circuits.
\newblock \emph{Distill}, 2021.
\newblock \doi{10.23915/distill.00024.006}.
\newblock https://distill.pub/2020/circuits/curve-circuits.

\bibitem[Conerly et~al.(2024)Conerly, Templeton, Bricken, Marcus, and Henighan]{conerly2024update}
Conerly, T., Templeton, A., Bricken, T., Marcus, J., and Henighan, T.
\newblock Update on how we train saes, 2024.

\bibitem[Cunningham et~al.(2023)Cunningham, Ewart, Riggs, Huben, and Sharkey]{cunningham2023sparse}
Cunningham, H., Ewart, A., Riggs, L., Huben, R., and Sharkey, L.
\newblock Sparse autoencoders find highly interpretable features in language models, 2023.

\bibitem[Elad(2010)]{elad2010sparse}
Elad, M.
\newblock \emph{Sparse and redundant representations: from theory to applications in signal and image processing}, volume~2.
\newblock Springer, 2010.

\bibitem[Elhage et~al.(2022)Elhage, Hume, Olsson, Schiefer, Henighan, Kravec, Hatfield-Dodds, Lasenby, Drain, Chen, et~al.]{elhage2022toy}
Elhage, N., Hume, T., Olsson, C., Schiefer, N., Henighan, T., Kravec, S., Hatfield-Dodds, Z., Lasenby, R., Drain, D., Chen, C., et~al.
\newblock Toy models of superposition.
\newblock \emph{arXiv preprint arXiv:2209.10652}, 2022.

\bibitem[Erhan et~al.(2009)Erhan, Bengio, Courville, and Vincent]{erhan2009visualizing}
Erhan, D., Bengio, Y., Courville, A., and Vincent, P.
\newblock Visualizing higher-layer features of a deep network.
\newblock \emph{University of Montreal}, 1341\penalty0 (3):\penalty0 1, 2009.

\bibitem[Marks et~al.(2024)Marks, Rager, Michaud, Belinkov, Bau, and Mueller]{marks2024sparse}
Marks, S., Rager, C., Michaud, E.~J., Belinkov, Y., Bau, D., and Mueller, A.
\newblock Sparse feature circuits: Discovering and editing interpretable causal graphs in language models.
\newblock \emph{arXiv preprint arXiv:2403.19647}, 2024.

\bibitem[Olah et~al.(2017)Olah, Mordvintsev, and Schubert]{olah2017feature}
Olah, C., Mordvintsev, A., and Schubert, L.
\newblock Feature visualization.
\newblock \emph{Distill}, 2017.
\newblock \doi{10.23915/distill.00007}.
\newblock https://distill.pub/2017/feature-visualization.

\bibitem[Olah et~al.(2020)Olah, Cammarata, Schubert, Goh, Petrov, and Carter]{olah2020an}
Olah, C., Cammarata, N., Schubert, L., Goh, G., Petrov, M., and Carter, S.
\newblock An overview of early vision in inceptionv1.
\newblock \emph{Distill}, 2020.
\newblock \doi{10.23915/distill.00024.002}.
\newblock https://distill.pub/2020/circuits/early-vision.

\bibitem[Olshausen \& Field(1997)Olshausen and Field]{olshausen1997sparse}
Olshausen, B.~A. and Field, D.~J.
\newblock Sparse coding with an overcomplete basis set: A strategy employed by v1?
\newblock \emph{Vision research}, 37\penalty0 (23):\penalty0 3311--3325, 1997.

\bibitem[Swee~Kiat(2021)]{torch-lucent}
Swee~Kiat, L.
\newblock Lucent, 2021.
\newblock URL \url{https://github.com/greentfrapp/lucent}.

\bibitem[Szegedy et~al.(2014)Szegedy, Liu, Jia, Sermanet, Reed, Anguelov, Erhan, Vanhoucke, and Rabinovich]{DBLP:journals/corr/SzegedyLJSRAEVR14}
Szegedy, C., Liu, W., Jia, Y., Sermanet, P., Reed, S.~E., Anguelov, D., Erhan, D., Vanhoucke, V., and Rabinovich, A.
\newblock Going deeper with convolutions.
\newblock \emph{CoRR}, abs/1409.4842, 2014.
\newblock URL \url{http://arxiv.org/abs/1409.4842}.

\bibitem[Voss et~al.(2021)Voss, Goh, Cammarata, Petrov, Schubert, and Olah]{voss2021branch}
Voss, C., Goh, G., Cammarata, N., Petrov, M., Schubert, L., and Olah, C.
\newblock Branch specialization.
\newblock \emph{Distill}, 2021.
\newblock \doi{10.23915/distill.00024.008}.
\newblock https://distill.pub/2020/circuits/branch-specialization.

\bibitem[Yun et~al.(2021)Yun, Chen, Olshausen, and LeCun]{yun2021transformer}
Yun, Z., Chen, Y., Olshausen, B.~A., and LeCun, Y.
\newblock Transformer visualization via dictionary learning: contextualized embedding as a linear superposition of transformer factors.
\newblock \emph{arXiv preprint arXiv:2103.15949}, 2021.

\end{thebibliography}
\bibliographystyle{icml2024}

\newpage
\appendix
\onecolumn
\section{Branch Specialisation of Learned Features}

After the first three convolution layers, InceptionV1 has a branched structure \cite{DBLP:journals/corr/SzegedyLJSRAEVR14}. Prior work has found that the early branches often specialise; for example, \texttt{mixed3b} seems specialize on curve and boundary detectors  \cite{voss2021branch}.

In the main portion of this paper, we trained sparse autoencoders on these branches as a principled way to save compute, with the intuition that the most computationally efficient way to represent a given feature is to utilise the most advantageous (i.e. the smallest) convolution size that can capture the feature. If this were true, the combination of neurons that represent a given feature would exist on the same branch.

\begin{figure*}[hbt!]
\begin{center}
\centerline{\includegraphics[width=\textwidth]{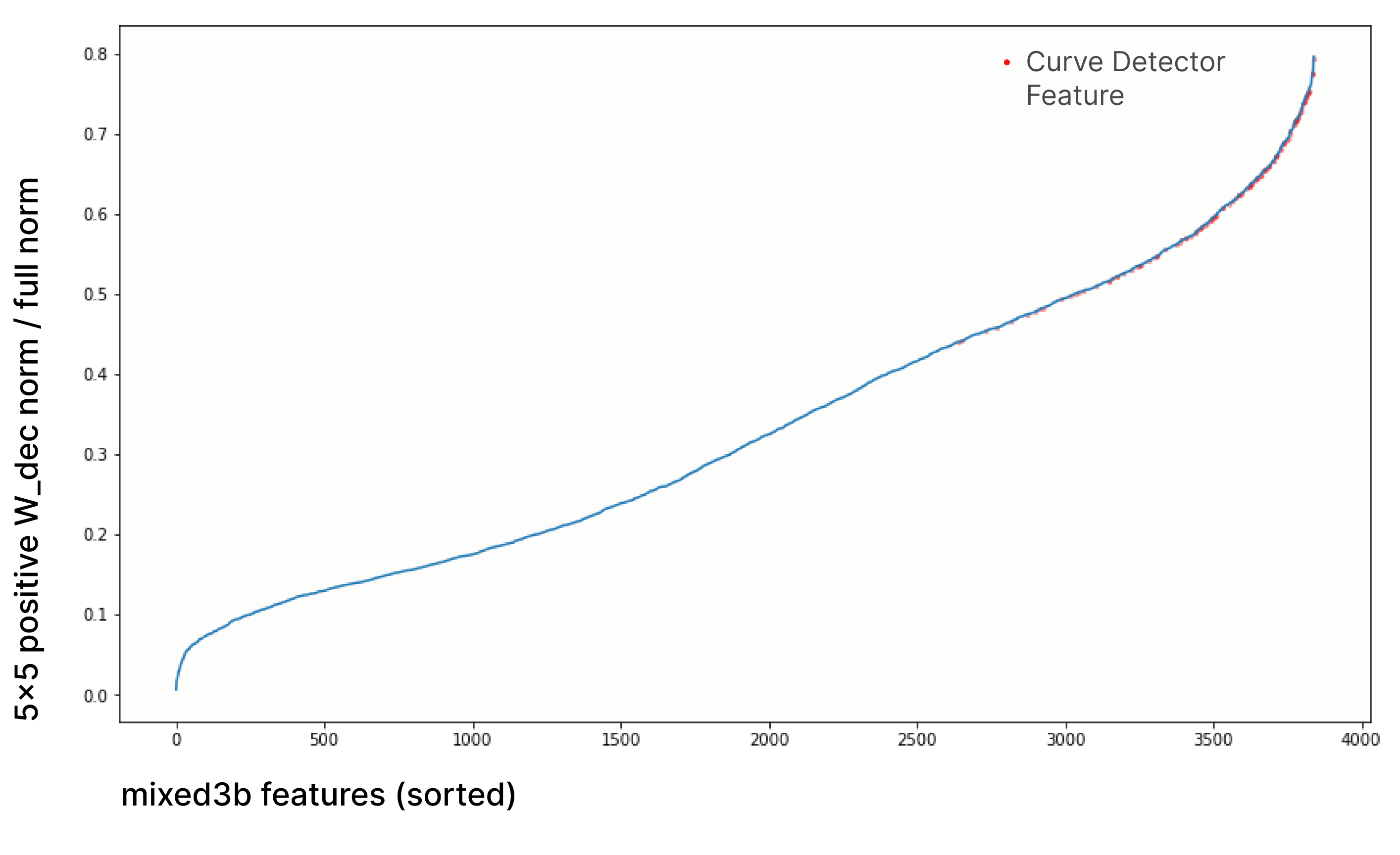}}
\caption{The extent to which each learned feature of a sparse autoencoder trained across all branches of \texttt{mixed3b}, is represented by neurons on the $5 \times 5$ branch. Negative weights were excluded when computing the norm. Curve detectors are amongst the features with the greatest branch specialisation.}
\label{branch-specialisation}
\end{center}
\end{figure*}

We now train sparse autoencoders on the entire \texttt{mixed3b} layer (ie. all branches jointly). We find that features are not uniformly distributed across branches. Some features are highly branch specific (as much as 80\% localized to \texttt{5x5}) and others are highly concentrated on other branches (nearly 0\% localized to \texttt{5x5}). On the other hand, while it's not clear where the boundaries for ``concentrated on a branch" should be (especially since \texttt{5x5} is smaller than other branches), but it seems likely that many features should be understood as spread across branches.

Interestingly, the features concentrated on \texttt{5x5} disproportionately tend to be curve detectors, consistent with \citet{voss2021branch}. 

What about other layers? In a preliminary SAE experiment on \texttt{mixed5b}, we find that a small fraction of features are very concentrated on the \texttt{5x5} branch, while most are more distributed. \texttt{mixed4e} preliminarily looks more similar to \texttt{mixed3b}. The nature of those branch-localized features is left to future work. Perhaps there is previously unknown branch specialization, which was previously obscured by superposition.

\begin{figure*}[hbt!]
\begin{center}
\centerline{\includegraphics[width=\textwidth]{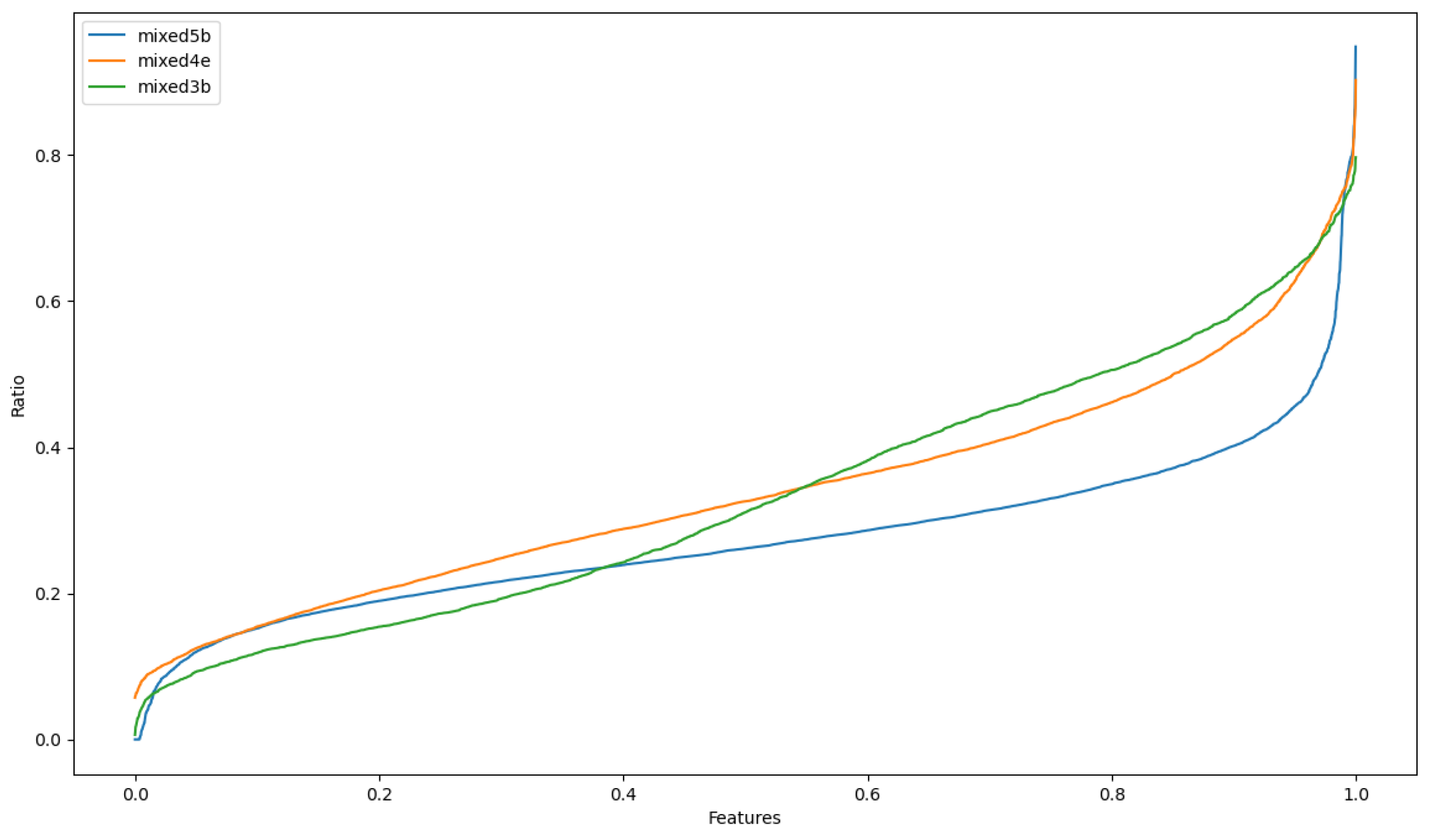}}
\caption{The extent to which each learned feature of sparse autoencoders trained across all branches of \texttt{mixed3b}, \texttt{mixed4e}, and \texttt{mixed5b} are represented by neurons on the $5 \times 5$ branch. Negative weights were excluded when computing the norm.}
\label{branch-specialisation}
\end{center}
\end{figure*}

\cref{branch-specialisation} shows how specialised each feature is to the $5\times5$ branch of \texttt{mixed3b}. No features are represented entirely on the $5\times5$ branch, with the most specialised feature having approximately $80\%$ of its positive weighted neurons on it.

It is worth emphasising that this doesn't change the validity of the features identified in this paper but instead means that branch-specific SAEs will have features still hidden by cross-branch superposition.

\end{document}